\documentclass[11pt,a4paper]{article}
\usepackage[hyperref]{naacl2021}
\usepackage{times}
\usepackage{latexsym}
\usepackage{booktabs}

\usepackage{microtype}
\usepackage{tabularx}

\usepackage{url}
\usepackage{calc}
\usepackage{xcolor}
\usepackage{graphicx}
\usepackage{multirow}

\usepackage{amsthm}
\usepackage{amssymb}
\usepackage{amsmath}
\usepackage{amsfonts}
\usepackage{color, colortbl}
\usepackage{adjustbox}
\usepackage{subcaption}
\usepackage{makecell}
\usepackage{arydshln}
\usepackage{bm}

\definecolor{Gray}{gray}{0.92}
\newcolumntype{Y}{>{\centering\arraybackslash}X}

\usepackage{todonotes}
\makeatletter
\newcommand*\iftodonotes{\if@todonotes@disabled\expandafter\@secondoftwo\else\expandafter\@firstoftwo\fi}
\makeatother

\title{Orthogonal Language and Task Adapters in \\ Zero-Shot Cross-Lingual Transfer}

\author{Marko Vidoni\textsuperscript{1}, Ivan Vuli\'{c}\textsuperscript{2}, and Goran Glava\v{s}\textsuperscript{3} \vspace{0.3em} \\
  \textsuperscript{1}Independent researcher, Berlin, Germany \\
  \textsuperscript{2}Language Technology Lab, University of Cambridge, UK \hspace{2mm} \vspace{0.3em} \\ 
  \textsuperscript{3}Data and Web Science Group, University of Mannheim, Germany \\
  \textsuperscript{1}{\tt marko.vidoni@gmail.com}, 
  \textsuperscript{2}{\tt iv250@cam.ac.uk} \\
  \textsuperscript{3}{\tt goran@informatik.uni-mannheim.de}
  
}

\date{}

\begin{document}
\maketitle
\begin{abstract}
Adapter modules, additional trainable parameters that enable efficient fine-tuning of pretrained transformers, have recently been used for language specialization of multilingual transformers, improving downstream zero-shot cross-lingual transfer. In this work, we propose \textit{orthogonal} language and task adapters (dubbed \textit{orthoadapters}) for cross-lingual transfer. They are trained to encode language- and task-specific information that is complementary (i.e., orthogonal) to the knowledge already stored in the pretrained transformer's parameters. Our zero-shot cross-lingual transfer experiments, involving three tasks (POS-tagging, NER, NLI) and a set of 10 diverse languages, \textbf{1)} point to the usefulness of orthoadapters in cross-lingual transfer, especially for the most complex NLI task, but also \textbf{2)} indicate that the optimal adapter configuration highly depends on the task and the target language. We hope that our work will motivate a wider investigation of usefulness of orthogonality constraints in language- and task-specific fine-tuning of pretrained transformers.  
\end{abstract}

\section{Introduction}
\label{sec:intro}

Multilingual representation spaces aim to capture meaning across language boundaries and that way (at least conceptually) enable cross-lingual transfer of task-specific NLP models from resource-rich languages with large annotated datasets to resource-lean languages without (m)any labeled instances \cite{joshi2020state}. When such a multilingual representation space is defined via parameters of a deep neural language encoder, e.g., a transformer network \cite{vaswani2017attention,devlin2019bert}, it can be used to transfer knowledge \textit{across tasks} as well as \textit{across languages} \cite[\textit{inter alia}]{pruksachatkun2020intermediate,ponti2020parameter,pfeiffer2020adapterhub,Phang:2020aacl} .
%

Massively multilingual transformers (MMTs), pretrained on large multilingual corpora via language modeling (LM) objectives \cite{devlin2019bert,Conneau:2020acl} have recently overthrown (static) cross-lingual word embedding spaces \cite{Ruder:2019jair,glavavs2019properly} as the state-of-the-art paradigm for zero-shot cross-lingual transfer in NLP. However, MMTs are constrained by the so-called \textit{curse of multilinguality}, a phenomenon where the quality of language-specific representations starts deteriorating when the number of pretraining languages exceeds the MMT's parameter capacity \cite{Conneau:2020acl}. Languages with smallest pretraining corpora are most affected: the largest transfer performance drops are observed with those target languages \cite{lauscher-etal-2020-zero,Wu:2020repl}. 

Additional LM training of a pretrained MMT on monolingual corpora of an underrepresented target language (i.e., target language \textit{LM-fine-tuning}) is a partial remedy towards satisfactory downstream transfer \cite{wang-etal-2020-extending,ponti-etal-2020-xcopa}. However, this approach does not increase the MMT capacity and, consequently, might deteriorate its representation quality for other languages. 

As a solution, \newcite{pfeiffer-etal-2020-mad} propose the MAD-X transfer framework: they freeze the original MMT parameters and train only a small number of additional parameters, the so-called \textit{adapter} modules \cite{rebuffi2017learning,pmlr-v97-houlsby19a} via language-specific LM-fine-tuning. This approach effectively \textbf{1)} increases the model capacity, and \textbf{2)} captures language-specific knowledge in the adapters.\footnote{In other words, there is a separate set of adapter parameters for each language with the added capacity reserved strictly for that language.} The next step is then task-specific fine-tuning on the source-language task data with a fixed source-language adapter (see Figure~\ref{fig:orthoadapters}): a separate set of \textit{task adapter} parameters is optimized to keep task-specific knowledge separate from language-specific knowledge stored in language adapters. Source language adapters are then replaced with target language adapters (with task adapters kept fixed), to make task predictions in the target language. The MAD-X framework, however, does not provide any mechanism that would prevent language and task adapters from \textit{capturing redundant information}, that is, from storing some knowledge already encoded in the MMT's parameters.

In this work, we advance the idea of augmenting MMT's knowledge through specialized adapter modules. We aim to maximize the injection of \textit{novel} information into both language- and task-specific adapter parameters, that is, we enforce the adapters to encode the information that \textit{complements} the knowledge encoded in MMT's pretrained parameters. To achieve this, we propose to learn \textit{orthogonal adapters} (or \textit{orthoadapters} for short). We augment the training objective\footnote{For a language adapter, the training task is masked language modeling on the monolingual corpus of that language.} with the \textit{orthogonality loss}: it forces the representations produced by the adapters to be \textit{orthogonal} to representations from the corresponding MMT layers. 

Our proof-of-concept zero-shot transfer experiments, encompassing cross-lingual transfer for three tasks -- POS-tagging, named entity recognition (NER), and natural language inference (XNLI) -- and 10 typologically diverse languages, render language-specific and task-specific orthoadapters viable mechanisms for improving downstream language transfer performance. However, we demonstrate that the optimal use of orthogonality is also largely task-dependent. We hope that our study will inspire a wider investigation of applicability and usefulness of orthogonality constraints in the context of both language-specific and task-specific fine-tuning of pretrained transformers. We will make the \textit{orthoadapters} code publicly available.




%










 
\section{Orthogonal Adapters}
\label{sec:orthoadapters}

\begin{figure}
    \centering
    \includegraphics[width=1.0\linewidth]{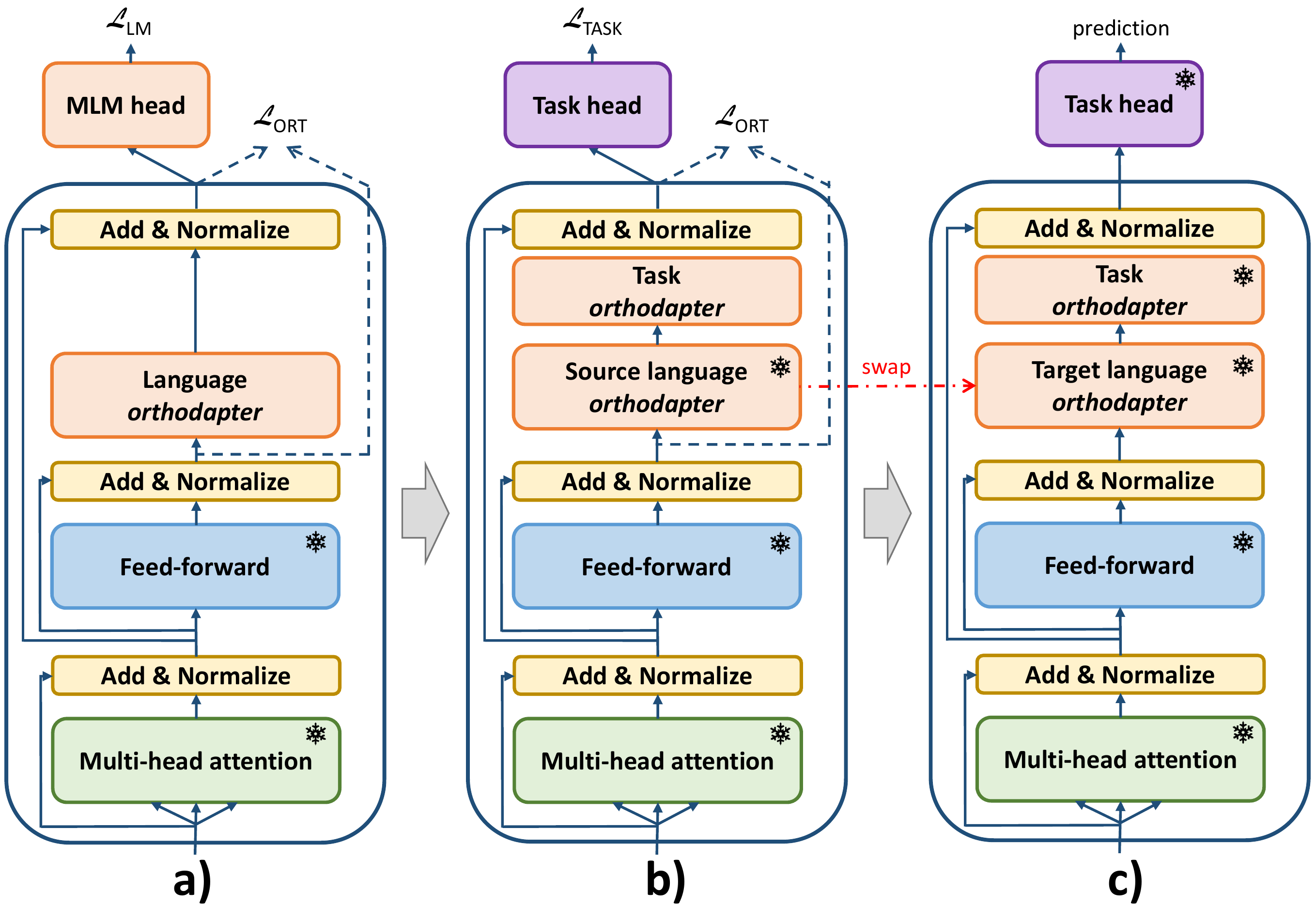}
    \caption{Orthogonal adapters for zero-shot cross-lingual transfer: \textbf{a)} Step 1: training language orthoadapters (independently for each language); \textbf{b)} Step 2: training the task orthoadapters on top of the (frozen) source-language orthoadapters on source-language data; \textbf{c)} Step 3: swaping the source-language orthoadapters with target language orthoadapters, allowing for task-specific target language inference. Snowflakes denote parameters which are kept frozen (i.e., not updated) in the respective step. For clarity, we show only a single transformer layer; the orthogonal adapter modules are used in all transformer layers of a pretrained MMT.}
    \label{fig:orthoadapters}
\end{figure}

Figure~\ref{fig:orthoadapters} provides an illustrative overview of our cross-lingual transfer framework for training and using language and task orthoadapters. 
%
%
We first train orthoadapters for each language: we inject a \textit{language adapter} into each transformer layer and train its parameters via masked language modeling (MLM) on the monolingual corpus of the respective language. We can augment the MLM objective with the {orthogonality loss}: it forces the adapter output to be orthogonal to the representations produced by the original transformer layer without the adapter (i.e., the multi-head attention (sub)layer coupled with the feed-forward layer). In the second step, we perform task-specific training in the source language. To this end, we inject the \textit{task adapter} to store task-specific knowledge. We can again aim to acquire knowledge that is absent from both the original MMT parameters by augmenting the task-specific training objective with the orthogonality loss. Keeping the language knowledge separate from task knowledge, we hope to impede negative interference and forgetting \cite{Hashimoto:2017emnlp,Autume:2019neurips,Wang:2020emnlp}.\footnote{Note that the task-specific fine-tuning procedure, in which we optimize task orthoadapters, is aware of the source language knowledge from the language orthoadapter.} Finally, as in the MAD-X framework, we swap the source-language adapter with corresponding target-language adapters for the task-specific inference on target-language data.   



\vspace{1.4mm}
\noindent \textbf{Adapter Architecture.} We adopt the well-performing and lightweight adapter configuration of \newcite{pfeiffer2020adapterfusion}, where only one adapter module is injected per transformer layer, after the feed-forward sublayer.\footnote{\newcite{pfeiffer2020adapterfusion} found this configuration to perform on a par with the configuration proposed by \newcite{pmlr-v97-houlsby19a}, who inject two adapter modules per transformer layer (the other one after the multi-head attention sublayer), while being more efficient to train.} The concrete adapter architecture we use is the variant of the so-called bottleneck adapter \cite{pfeiffer2020adapterfusion}:
\begin{equation}
    \mathbf{x}_a = \mathit{Adapt}(\mathbf{x}_h, \mathbf{x}_r) = g\left(\mathbf{\mathbf{x}_h} \mathbf{W}_d\right)\mathbf{W}_u + \mathbf{x}_r
    \label{eq:adapter}
\end{equation}
\noindent where $\mathbf{x}_h$ and $\mathbf{x}_r$ are the hidden state
and residual representation of the transformer layer, respectively. The parameter matrix $\mathbf{W}_d \in \mathbb{R}^{H \times d}$ down-projects (i.e., compresses) hidden representations to the \textit{adapter size} $d < H$, and $\mathbf{W}_u \in \mathbb{R}^{d \times H}$ up-projects the activated down-projections back to the transformer's hidden size $H$; $g$ is the non-linear ReLU activation \cite{Nair:2010icml}. 

\vspace{1.4mm}
\noindent \textbf{Orthogonality Loss.} 
Adapters allow for computationally more efficient fine-tuning \cite{pmlr-v97-houlsby19a}, while simultaneously offering task performance comparable to that of standard \textit{full fine-tuning} of all transformer parameters \cite{pfeiffer2020adapterfusion}. However, there is currently no mechanism in adapter-based fine-tuning that would explicitly prevent adapter parameters from learning redundant information which is already captured by the pretrained model. 
Inspired by the idea of orthogonal text representations introduced in the context of multi-task learning with shared and task-specific (i.e., private) parameters \cite{Romera:2012aistats,liu-etal-2017-adversarial}, we introduce an additional orthogonality loss in adapter-based fine-tuning. It \textit{explicitly} forces the adapters to dedicate their capacity to \textit{new} knowledge, which should be \textit{complementary} (i.e., non-redundant) to the knowledge already encoded in existing transformer's parameters.

In particular, we can augment the training objective in each of the two training phases (i.e., when training both language adapters and task adapters) with an auxiliary orthogonality loss. Let $\mathbf{x}^{(i,j)}_h$ denote the hidden representation in the $i$-th layer of the MMT for the $j$-th token in the sequence, input to the adapter. Let $\mathbf{x}^{(i,j)}_a$ be the corresponding output of the same adapter for the same token, as given in Eq.~\eqref{eq:adapter}. The orthogonality loss of the $j$-th token in the $i$-th MMT layer is then simply the square of the cosine similarity between $\mathbf{x}^{(i,j)}_h$ and $\mathbf{x}^{(i,j)}_a$; we then derive the overall orthogonality loss of the model by averaging token-level losses in each layer and then summing layer-level losses:  
\begin{equation}
\mathcal{L}_\mathit{ORT} = \sum^{N}_{i = 1}{\frac{1}{T}\sum_{j=1}^{T} {\cos\left(\mathbf{x}^{(i,j)}_h, \mathbf{x}^{(i,j)}_a\right)^2}}
\end{equation}
where $T$ is the maximal length of the input token sequence and $N$ is the number of MMT's layers. 

\vspace{1.4mm}
\noindent \textbf{Two-Step Orthoadapter Training.} In the first step (see part \textbf{a} of Figure~\ref{fig:orthoadapters}), we train language orthoadapters, independently for each language. This way we aim to extend the language knowledge captured by the pretrained MMT. As in MAD-X \cite{pfeiffer-etal-2020-mad}, we use masked LM-ing (cross-entropy loss) as the main training objective $\mathcal{L}_\mathit{MLM}$. We then alternately update the parameters of language orthoadapters, first by minimizing $\mathcal{L}_\mathit{MLM}$ and then by minimizing $\mathcal{L}_\mathit{ORT}$.\footnote{We use two independent Adam optimizers \cite{kingma2015adam}, one for each loss. We also experimented with minimizing the joint loss $\mathcal{L}_\mathit{MLM} + \lambda \cdot \mathcal{L}_\mathit{ORT}$ but this generally yielded poorer performance although we tried a wide range of values for the weight $\lambda$.} 

In the second step (part \textbf{b} in Figure~\ref{fig:orthoadapters}), the goal is to maximize the amount of novel information useful for a concrete downstream task: we train the task orthoadapters on the task-specific training data (POS, NER, XNLI) by alternately minimizing \textbf{1)} the task-specific objective $\mathcal{L}_\mathit{TASK}$\footnote{Cross-entropy loss for the whole sequence for NLI; sum of token-level cross-entropy losses for POS and NER.} and \textbf{2)} the orthogonality loss $\mathcal{L}_\mathit{ORT}$. Note, however, that in this case $\textbf{x}_h$ is, in each transformer layer, first adapted by the source language adapter and then by the task adapter, and $\textbf{x}_a$ is the output of the task adapter. 

\vspace{1.4mm}
\noindent \textbf{Zero-Shot Cross-Lingual Transfer} then proceeds in the same vein as in the MAD-X framework \cite{pfeiffer-etal-2020-mad}. The source-to-target transfer is conducted by simply replacing the source language (ortho)adapter with the target language (ortho)adapter while relying on exactly the same task adapter fine-tuned with the labeled source language data, stacked on top of the language adapters (see part \textbf{c} of Figure \ref{fig:orthoadapters}). We refer the reader to the original work \cite{pfeiffer-etal-2020-mad,pfeiffer2020adapterhub} for further technical details.

\begin{table*}[t]
\def\arraystretch{0.99}
\centering
{\footnotesize
\begin{tabularx}{\linewidth}{l YY Y Y}
\toprule 
{\bf Language} & {Family} & {Type} & {ISO 639} & {Tasks}\\
\cmidrule(lr){2-5}
{English} & {IE: Germanic} & {fusional} & {\textsc{en}} & {XNLI, UD-POS, NER}\\
\hdashline
\rowcolor{Gray}
{Arabic} & {Semitic} & {introflexive} & {\textsc{ar}} & {XNLI, UD-POS, NER}\\
{Estonian} & {Uralic: Finnic} & {agglutinative} & {\textsc{et}} & {UD-POS, NER}\\
\rowcolor{Gray}
{Hindi} & {IE: Indo-Aryan} & {fusional} & {\textsc{hi}} & {XNLI, UD-POS, NER}\\
{Ilocano} & {Austronesian} & {agglutinative} & {\textsc{ilo}} & {NER}\\
\rowcolor{Gray}
{Meadow Mari} & {Uralic: Mari} & {agglutinative} & {\textsc{mhr}} & {NER} \\
{Quechua} & {Quechuan} & {agglutinative} & {\textsc{qu}} & {NER}\\
\rowcolor{Gray}
{Kiswahili} & {Niger-Congo: Bantu} & {agglutinative} & {\textsc{sw}} & {XNLI, NER}\\
{Turkish} & {Turkic} & {agglutinative} & {\textsc{tr}} & {XNLI, UD-POS, NER} \\
\rowcolor{Gray}
{Mandarin Chinese} & {Sino-Tibetan} & {isolating} & {\textsc{zh}} & {XNLI, UD-POS, NER} \\
\bottomrule
\end{tabularx}
}
\vspace{-1.5mm}
\caption{Target languages used in the main experiments along with their corresponding language family (IE=Indo-European), morphological type, and ISO 639-1 code (or ISO 639-2 for Ilocano; or ISO 639-3 for Meadow Mari). We use English (\textsc{en}) as the source language in all experiments. \textsc{en} is a fusional language (IE: Germanic).}
\label{tab:langs}
\vspace{-1.5mm}
\end{table*}

\section{Experimental Setup}
\label{sec:exp}
\noindent \textbf{Model Configurations.} The decomposition into two adapter types in the two-step procedure (Figure~\ref{fig:orthoadapters}) allows us \textbf{1)} to use language orthoadapters (\textsc{l-ort}) instead of \textit{regular non-orthogonal} language adapters (\textsc{l-noo}); and/or \textbf{2)} to replace non-orthogonal task adapters (\textsc{t-noo}) with task orthoadapters (\textsc{t-ort}). These choices give rise to four different model variants, where the \textsc{l-noo+t-noo} variant is the baseline MAD-X variant.

We also test the usefulness of task orthoadapters in a standard ``non-MAD-X'' setup, i.e., without dedicated language adapters: \textsc{t-ort} variants are compared to \textsc{t-noo} variants, and also to standard full fine-tuning of the whole MMT (\textsc{full-ft}) with in-task labelled data, which is computationally more intensive than adapter-based fine-tuning.  

\vspace{1.4mm}
\noindent \textbf{Evaluation Tasks and Data.}
 We evaluate all model variants on standard cross-lingual transfer tasks, relying on established evaluation benchmarks: \textbf{1)} sentence-pair classification on XNLI \cite{Conneau2018xnli}); \textbf{2)} cross-lingual named entity recognition (NER) on the WikiANN dataset \cite{Pan2017};\footnote{As prior work \cite{pfeiffer-etal-2020-mad,Hu:2020arxiv}, we use the data splits provided by \newcite{Rahimi:2019acl}.} \textbf{3)} part-of-speech tagging with universal POS tags from the Universal Depenedencies \cite{nivre2018universal} (UD-POS).
 
 In all experiments we rely on the pretrained multilingual XLM-R (Base) model \cite{Conneau:2020acl}, which showed state-of-the-art zero-shot performance in a recent comparative empirical study of \newcite{Hu:2020arxiv}, and even stronger results when combined with the adapter-based MAD-X framework \cite{pfeiffer-etal-2020-mad}. We always treat English (\textsc{en}) as our (resource-rich) source language: the pretrained XLM-R model is fine-tuned on the English task data, and then evaluated in the zero-shot setting on different target languages. For completeness, we also report the results on the English test data, i.e., without any transfer.
 
\vspace{1.4mm}
\noindent \textbf{Target Languages.} The selection of target languages has been guided by several (sometimes clashing) criteria: \textbf{C1)} typological diversity; \textbf{C2)} availability in the standard evaluation benchmarks; \textbf{C3)} computational tractability; \textbf{C4)} evaluation also on truly low-resource languages. Given that the main computational bottleneck is MLM-ing for learning language adapters, we have started from the subset of languages represented in our evaluation datasets (C2) for which pretrained language adapters (regular, non-orthogonal) are already available online (C3) \cite{pfeiffer2020adapterhub}. The final list of target languages, available in Table~\ref{tab:langs} with their corresponding language codes, comprises 10 languages from 5 geographical macro-areas \cite{wals,ponti-etal-2020-xcopa} representing 8 distinct language families and covering at least one language for each broad morphosyntactic language type (isolating, introflexive, fusional, agglutinative), satisfying C1. Finally, in NER evaluations
we include three truly low-resource languages: Quechua, Ilocano, and Meadow Mari (C4).


\vspace{1.4mm}
\noindent \textbf{Training and Evaluation: Technical Details.} We rely on the AdapterHub library \cite{pfeiffer2020adapterhub} built on top of the Transformers library \cite{Wolf:2020emnlp} in all experiments based on the MAD-X framework. For training language and task adapters we follow the suggestions from prior work \cite{pmlr-v97-houlsby19a,pfeiffer-etal-2020-mad}. For orthogonal language adapters, we conduct MLM-ing on the Wikipedia data of each language. For learning task orthoadapters, we rely on the standard training portions of our task data in English.\footnote{Unlike prior work that effectively violates the zero-shot assumption by doing model selection using development data in the target language \cite{Conneau:2020acl,Keung:2020emnlp}, we select task (ortho)adapters solely based on the performance on the source language (i.e, English) dev set.} We provide the full details of our training and fine-tuning procedures (including the details on hyperparameter search), in the Appendix~A.

\begin{table*}[!t]
    \centering
    {\small
    \begin{tabularx}{\textwidth}{l YYYYYY Y}
    \toprule
    {\bf Variant} & {\textsc{en}} & {\textsc{ar}} & {\textsc{hi}} & {\textsc{sw}} & {\textsc{tr}} & {\textsc{zh}} & {\bf AVG\textit{z}} \\
    \cmidrule(lr){2-7}
    {\textsc{full-ft}} & {83.67} & {72.01} & {68.64} & {63.77} & {71.75} & {73.11} & {69.86} \\
    {\textsc{t-noo}} & {84.05} & {69.51} & {68.26} & {64.48} & {71.73} & {72.25} & {69.25} \\
    \rowcolor{Gray}
    {\textsc{t-ort}} & {84.25} & {69.97} & {68.84} & {63.35} & {70.85} & {71.95} & {68.99} \\
    \hdashline
    {\textsc{l-noo+t-noo}} & {84.59} & {71.17} & {70.61} & {67.68} & {71.75} & {72.29} & {70.70} \\
    \rowcolor{Gray}
    {\textsc{l-noo+t-ort}} & {84.79} & {68.88} & {69.71} & {66.34} & {70.47} & {71.49} & {69.38} \\
    \rowcolor{Gray}
    {\textsc{l-ort+t-noo}} & {84.73} & {72.25} & {69.28} & {69.10} & {72.89} & {73.61} & {71.43} \\
    \rowcolor{Gray}
    {\textsc{l-ort+t-ort}} & {84.35} & {69.73} & {68.56} & {67.92} & {71.23} & {71.53} & {69.79} \\
    \bottomrule
    \end{tabularx}
    }%
    \vspace{-1.5mm}
    \caption{Accuracy scores ($\times$100$\%$) of zero-shot transfer for the natural language inference task on the XNLI dataset. See \S\ref{sec:exp} for the descriptions of different model variants. \textsc{en} is the source language in all experiments. The scores in the AVG\textit{z} column denote the average performance of zero-shot transfer (i.e., without English results).}
    \label{tab:xnli}
    \vspace{-1mm}
\end{table*}


\begin{table*}[!t]
    \centering
    {\small
    \begin{tabularx}{\textwidth}{l YYYYYY c}
    \toprule
    {\bf Variant} & {\textsc{en}} & {\textsc{ar}} & {\textsc{et}} & {\textsc{hi}} & {\textsc{tr}} & {\textsc{zh}} & {\bf AVG\textit{z}}\\
    \cmidrule(lr){2-7}
    {\textsc{full-ft}} & {95.94} & {66.42} & {84.68} & {70.38} & {74.01} & {35.59} & {66.22} \\
    {\textsc{t-noo}} & {95.59} & {64.35} & {84.43} & {71.06} & {72.68} & {31.47} & {64.80} \\
    \rowcolor{Gray}
    {\textsc{t-ort}} & {95.65} & {65.28} & {85.17} & {70.42} & {72.93} & {40.03} & {66.77} \\
    \hdashline
    {\textsc{l-noo+t-noo}} & {95.59} & {65.77} & {85.81} & {69.62} & {74.17} & {19.25} & {62.92} \\
    \rowcolor{Gray}
    {\textsc{l-noo+t-ort}} & {95.66} & {67.15} & {85.67} & {71.57} & {74.08} & {31.68} & {66.03} \\
    \rowcolor{Gray}
    {\textsc{l-ort+t-noo}} & {95.63} & {66.62} & {84.26} & {68.93} & {72.02} & {29.50} & {64.27} \\
    \rowcolor{Gray}
    {\textsc{l-ort+t-ort}} & {95.63} & {67.40} & {84.22} & {67.26} & {71.21} & {31.79} & {64.38} \\
    \bottomrule
    \end{tabularx}
    }%
    \vspace{-1.5mm}
    \caption{$F_1$ scores ($\times$100$\%$) of zero-shot transfer in the UD-POS task.}
    \label{tab:udpos}
    \vspace{-1mm}
\end{table*}

\begin{table*}[!t]
    \centering
    {\small
    \begin{tabularx}{\textwidth}{l YYYYYYYYYY Y}
    \toprule
    {\bf Variant} & {\textsc{en}} & {\textsc{ar}} & {\textsc{et}} & {\textsc{hi}} & {\textsc{ilo}} & {\textsc{mhr}} & {\textsc{qu}} & {\textsc{sw}} & {\textsc{tr}} & {\textsc{zh}} & {\bf AVG\textit{z}}\\
    \cmidrule(lr){2-11}
    {\textsc{full-ft}} & {82.98} & {30.85} & {65.63} & {58.87} & {62.04} & {39.83} & {60.43} & {61.26} & {65.98} & {8.77} & {50.41} \\
    {\textsc{t-noo}} & {83.41} & {46.19} & {70.86} & {67.14} & {60.75} & {39.33} & {58.01} & {60.79} & {72.60} & {23.07} & {55.42} \\
    \rowcolor{Gray}
    {\textsc{t-ort}} & {82.87} & {46.71} & {70.27} & {66.16} & {59.03} & {45.30} & {58.72} & {61.26} & {72.93} & {19.69} & {55.56} \\
    \hdashline
    {\textsc{l-noo+t-noo}} & {82.33} & {41.29} & {75.04} & {64.84} & {63.11} & {54.55} & {70.64} & {71.90} & {71.63} & {16.68} & {58.85} \\
    \rowcolor{Gray}
    {\textsc{l-noo+t-ort}} & {82.86} & {38.74} & {74.48} & {64.21} & {69.72} & {53.01} & {61.02} & {72.78} & {70.65} & {14.41} & {57.67} \\
    \rowcolor{Gray}
    {\textsc{l-ort+t-noo}} & {82.44} & {40.62} & {74.27} & {66.74} & {77.21} & {50.78} & {65.31} & {72.98} & {70.80} & {12.16} & {58.99} \\
    \rowcolor{Gray}
    {\textsc{l-ort+t-ort}} & {82.63} & {35.82} & {72.81} & {61.83} & {70.64} & {52.31} & {64.98} & {73.84} & {69.41} & {12.25} & {57.10} \\
    \bottomrule
    \end{tabularx}
    }%
    \vspace{-1.5mm}
    \caption{$F_1$ scores ($\times$100$\%$) of zero-shot transfer in the NER task on the WikiAnn dataset.}
    \label{tab:ner}
    \vspace{-0.75em}
\end{table*}

\section{Results and Discussion}
\label{sec:results}
The results of zero-shot transfer for the three evaluation tasks are summarized in Table~\ref{tab:xnli} (for the XNLI task), Table~\ref{tab:udpos} (UD-POS), and Table~\ref{tab:ner} (NER). For all three tasks, besides scores per language, we also include the average results of \textit{z}ero-shot transfer experiments, i.e., excluding English test scores as the non-transfer experiment (the AVG\textit{z} column). A starting observation based on the comparison with the full fine-tuning variant (\textsc{full-ft}) confirms findings from previous work \cite{pfeiffer2020adapterfusion}, further validating the use of the more efficient adapter-based fine-tuning: the scores with adapter-based variants are on a par with or even higher than the scores reported with \textsc{full-ft} across the board.

\vspace{1.4mm}
\noindent \textbf{Regular vs Orthogonal Language Adapters.} Our initial set of result indicates that the usefulness of our orthogonal language adapters (\textsc{L-ORT} variants) does depend on the task at hand and its complexity. As an encouraging finding, we observe consistent gains in cross-lingual NLI, which is arguably the most complex (reasoning) task in our evaluation and, unlike UD-POS and NER, requires successful modeling of complex semantic compositionality in a (target) language. The gains (at least +1 accuracy point) are observed for 4 out of 5 target languages when doing zero-shot transfer with the \textsc{l-ort+t-noo} variant. This variant also yields highest average transfer performance, and slight (but statistically insignificant) gains on English NLI. However, the picture is less clear for UD-POS and NER: \textsc{l-ort+t-noo} does have a slight edge over the baseline fully non-orthogonal variant (\textsc{l-noo+t-noo}) in UD-POS, but this seems to mostly stem from large gains in Chinese. In a similar vein, while the \textsc{l-ort+t-noo} variant is the best performing variant in NER experiments on average, the gains over the baseline \textsc{l-noo+t-noo} are slight, and inconsistent across languages (e.g., large gains on \textsc{ilo}, improvements on \textsc{ar} and \textsc{sw}, but some decrease on \textsc{qu} and \textsc{mhr}).

We speculate that this is mostly due to the nature and complexity of the task at hand. In order to perform cross-lingual transfer for language inference, the underlying MMT must capture and leverage more language-specific nuances than for sequence labeling tasks such as POS-tagging or NER. By enforcing the capture of non-redundant information in the additional language-specific adapter modules, we allow the model to store additional and, more importantly, \textit{novel} target language information. While the same information is available also for NER and POS tagging, these tasks require 'shallower' language-specific knowledge \cite{lauscher-etal-2020-zero} and this is why more complex target language-specific knowledge captured in orthoadapters (compared to regular non-orthogonal language adapters) does not make a difference. 

\vspace{1.4mm}
\noindent \textbf{Orthogonal Task Adapters}, on the other hand, display a different behavior, but we can again largely relate it to the properties of the evaluation tasks. First, the use of task orthoadapters seems detrimental across the board in the MAD-X setup for XNLI (compare \textsc{l-noo+t-ort} vs.\,\textsc{l-noo+t-noo} as well as \textsc{l-ort+t-ort} vs.\,\textsc{l-ort+t-noo} in Table \ref{tab:xnli}), and also yields no real benefits in the simpler setup with no language adapters (\textsc{t-ort} vs.\,\textsc{t-noo}). We hypothesize that the two main objectives -- (i) MLM for the original MMT pretraining and language adapter training, and (ii) cross-entropy loss for the whole sequence for NLI -- are structurally too different for the orthogonality loss to capture any additional task-related information. In fact, we speculate that the orthogonal loss might have emphasized this discrepancy between the objectives and actually even hurt the final performance.  

However, task orthoadapters seem to be useful for UD-POS transfer, with substantial gains reported on 3/5 target languages -- Arabic, Chinese, Hindi, all of which have non-Latin scripts, while there is no change in performance for Estonian and Turkish. Combining task orthoadapters with language orthoadapters, however, does deteriorate the performance. The overall trend is even more complex in NER transfer: while there are clear hints that using orthoadapters is useful for some languages and some model variants, there is still a substantial variance in the results. We partially attribute it to the documented volatility of the WikiAnn evaluation set, especially for low-resource languages \cite{pfeiffer-etal-2020-mad}. 

In general, our results suggest that explicitly controlling for the information that gets captured in the adapter modules can have a positive impact on cross-lingual transfer via MMTs. The optimal use of the orthogonality loss, however, seems to be largely target-language- and task-dependent, warranting further investigations in future work.

\section{Conclusion and Future Work}
\label{sec:conclusion}

We have investigated how orthogonality constraints impact downstream performance of zero-shot cross-lingual transfer via massively multilingual transformers (e.g., XLM-R) for three standard tasks: NLI, POS, and NER. Relying on the standard adapter-based transfer techniques, we have introduced the idea of orthogonal language and task adapters (or orthoadapters): we explicitly enforce the information stored in the parameters of language and task adapters to be orthogonal to the information already stored in the pretrained MMT. With such an orthogonality mechanism in place we should be able to encode novel rather than redundant information. Our results have demonstrated the validity of orthoadapters, especially in the most complex XNLI task, although the optimal adapter setup seems language- and task-dependent.

Our work has pointed to the importance of enforcing and controlling what information gets stored in the adapter modules for improved cross-lingual transfer, but it has only scratched the surface. In future work, we will investigate more sophisticated orthogonality losses \cite{Bousmalis:2016neurips,liu-etal-2017-adversarial} and techniques such as gradual adapter unfreezing \cite{Howard:2018acl,Peters:2019repl}. We will also explore the usefulness of orthoadapters in other tasks as well as for domain adaption \cite{Ruckle:2020emnlp}.


\bibliography{references}
\bibliographystyle{acl_natbib}

\appendix


\section{Training Details}




\subsection{Task Training Details}
We processed the data for all tasks using the preprocessing pipeline provided with the XTREME benchmark \cite{Hu:2020arxiv}.\footnote{\url{https://github.com/google-research/xtreme}}


\vspace{1.4mm}
\noindent \textbf{XNLI.} In NLI training (i.e., for XNLI transfer) were trained for 30 epochs with the batch size of 32. Maximum sequence length was 128 input tokens. Gradient norms were clipped to 1.0.

\vspace{1.4mm}
\noindent \textbf{UD-POS.} We trained for 50 epochs with batch size of 16. Maximum sequence length was 128 input tokens. Gradient norms were clipped to 1.0.

\vspace{1.4mm}
\noindent \textbf{NER.} We trained for 100 epochs with the batch size of 16. Maximum sequence length was 128 input tokens. Gradient norms were clipped to 1.0.



\subsection{Experimental Setup without Language Adapters.}
For the ``non-MAD-X'' experimental setup (i.e., the setup without language adapters, see \S\ref{sec:exp}), we relied on our own implementation of the adapter module. The bottleneck size for the task adapter was set to $d = 64$. For (X)NLI, we searched the following learning rate grid: $[2e-5, 5e-5, 7e-5]$; for UD-POS and WikiAnn the corresponding learning rate grid was $[5e-5, 1e-4]$. For task orthoadapters, we searched the following additional learning rate grid for the orthogonal loss optimizer: $ [1e-6, 1e-7, 1e-8]$. 

\subsection{Full Experimental Setup}
For the more complex multi-adapter setup based on the MAD-X framework (i.e., with both language and task adapters), we utilized the \textit{Adapter-Transformers} library and the underlying \textit{AdapterHub} service \cite{pfeiffer2020adapterhub}. 


\vspace{1.4mm}
\noindent \textbf{Task Adapters and Orthoadapters.}
We followed the recommendation from the original paper \cite{pfeiffer-etal-2020-mad}. We utilized the \textit{Pfeiffer} configuration found in the Adapter-Transformers library with the adapter dimensionality of 48. Due to the computational constraints, the learning rate grid search took into account best settings observed in the baseline experiments. For XNLI our main learning rate was set at a well-performing $5e-5$. For UD-POS and WikiAnn, due to more instability, we tested the learning rate grid of $[5e-5, 1e-4]$. For the task orthoadapters, we used the same learning rate for the orthogonality loss as for the non-MAD-X setup: $[1e-6, 1e-7, 1e-8]$.


\vspace{1.4mm}
\noindent \textbf{Regular Language Adapters.} We utilized the pretrained language adapters readily available via the AdapterHub service \cite{pfeiffer2020adapterhub}. These language adapters have the dimensionality of 384. They were trained (while the rest of the model was frozen) by executing the MLM-ing for 250.000 iterations on the Wikipedia data in the target language.
 
\vspace{1.4mm}
\noindent \textbf{Orthogonal Language Adapters.} We started from the MLM-ing training script for training language adapters provided by the Adapter-Transformers library and trained language orthoadapters on the Wikipedia data, relying on the setup of AdapterHub's regular language adapters (dimensionality 384, 250,000 iterations). Due to computational constraints we reduced the maximum sequence length of the input to 128 tokens, while the batch size was 8. Finally, for the main optimizer and orthogonality loss optimizer we used the learning rates of $ 1e-4$ and $1e-7$, respectively.


\label{sec:appendix}

\end{document}